# SDN Flow Entry Management Using Reinforcement Learning

TING-YU MU, Western Michigan University, Kalamazoo MI 49008 U.S.A
ALA AL-FUQAHA, Western Michigan University, Kalamazoo MI 49008 U.S.A
KHALED SHUAIB, United Arab Emirates University
FARAG M. SALLABI, United Arab Emirates University
JUNAID QADIR, Information Technology University, Lahore Pakistan

Modern information technology services largely depend on cloud infrastructures to provide their services. These cloud infrastructures are built on top of datacenter networks (DCNs) constructed with high-speed links, fast switching gear, and redundancy to offer better flexibility and resiliency. In this environment, network traffic includes long-lived (elephant) and short-lived (mice) flows with partitioned/aggregated traffic patterns. Although SDN-based approaches can efficiently allocate networking resources for such flows, the overhead due to network reconfiguration can be significant. With limited capacity of Ternary Content-Addressable Memory (TCAM) deployed in an OpenFlow enabled switch, it is crucial to determine which forwarding rules should remain in the flow table, and which rules should be processed by the SDN controller in case of a table-miss on the SDN switch. This is needed in order to obtain the flow entries that satisfy the goal of reducing the long-term control plane overhead introduced between the controller and the switches. To achieve this goal, we propose a machine learning technique that utilizes two variations of reinforcement learning (RL) algorithms—the first of which is traditional reinforcement learning algorithm based while the other is deep reinforcement learning based. Emulation results using the RL algorithm show around 60% improvement in reducing the long-term control plane overhead, and around 14% improvement in the table-hit ratio compared to the Multiple Bloom Filters (MBF) method given a fixed size flow table of 4KB.

## 1 INTRODUCTION

The rapid growth of big data processing and the demand for massive-scale datacenters has increased the need for more efficient and intelligent network management systems, which has motivated an evolution of networking architecture towards Software-Defined Networking (SDN) [1], [3]. SDN is a popular networking paradigm of programmable networks that decouples the packet forwarding mechanism (data plane) from the control decisions (control plane). The goal of SDN is to provide an open programmable interface allowing the development of applications that dynamically control and manage connectivity among network elements. This enables the network to be more "application-aware," and "network-aware."

Currently, OpenFlow [2], [3] is the dominating protocol used by controllers and switches in an SDN to exchange information. OpenFlow allows a switch to notify the controller of an incoming packet for which no forwarding rules can be found in the flow table. Similarly, a controller can send control messages to a switch requesting it to add a new or modify an existing forwarding rule in the flow table to serve the incoming packets. This information exchange process is also referred to as the network control overhead in this paper. This process consumes networking resources and introduces latency to the packet forwarding process.

In order to analyze traffic patterns in datacenters, traffic flows are further classified as either throughput-dependent (elephant flows) or short-lived (mice flows). From network traffic perspective, transactional applications (e.g., web searches, social networking) are usually executed by several endpoints that generate several small traffic flows requiring timely delivery to satisfy the end users' delay requirements. On the other hand, bulk transfers (e.g., MapReduce, and virtual machine migrations) require minimal packet delivery delays while utilizing a high and sustained network bandwidth. Previous studies on live datacenter traffic patterns [4], [5] indicate that the number of elephant flows is less than 10% of all flows flowing within the DCN, but the payloads carried by such flows account for more than 80% of the entire traffic volume. In addition, due to the short-lived lifespan of the mice flows, they may not have any significant impact on the DCNs if considered individually. However, mice flows represent 90% of the entire flows [4], [5], and improper handling of mice flows can be a major performance hindrance.

The co-existence of such elephant and mice traffic patterns presents a challenge in DCNs as the large number of mice flows can cause the elephant forwarding rules to be evicted from the flow tables prematurely, especially when a burst of mice packets arrive at a switch in the middle of an ongoing elephant flow [6]. Given that every network switch has a limited amount of memory space for storing forwarding rules (i.e., flow entries), it is highly likely that the switch will not have enough space to accommodate additional flow entries for packets that require new forwarding rules. This process can ultimately lead to packet drops and/or significant packet delays.

Due to the dynamic and heterogeneous nature of DCNs (especially in terms of their offered load and available resources), it is difficult to formulate a mathematical model designed to properly manage the forwarding rules and adapt to the monitored pattern of elephant and mice flows. To address this issue, we utilize Reinforcement Learning (RL) to automatically search for a near optimal set of forwarding rules that minimize the long-term control plane overhead. The goal is to minimize the long-term table-miss events and packet delivery latencies given the Ternary Content-Addressable Memory (TCAM) size of the SDN switches. We are motivated to use RL by its impressive capabilities of adapting to unknown operating environments and improving its decision policies using reinforcement obtained through previous interactions with the environment. We believe that RL is well suited to the task of automatic self-tuning as it allows the learning agent to explore the flow configurations space through the feedback supplied by the environment based on the action performed by the agent with the goal to establish the policy that maximizes the long-term reward. We also note that RL is better suited than traditional optimization techniques (e.g., discrete optimization techniques such as integer linear programming) since RL is adaptive and learns in dynamic environments unlike traditional optimization techniques that only compute the optimum for a given snapshot.

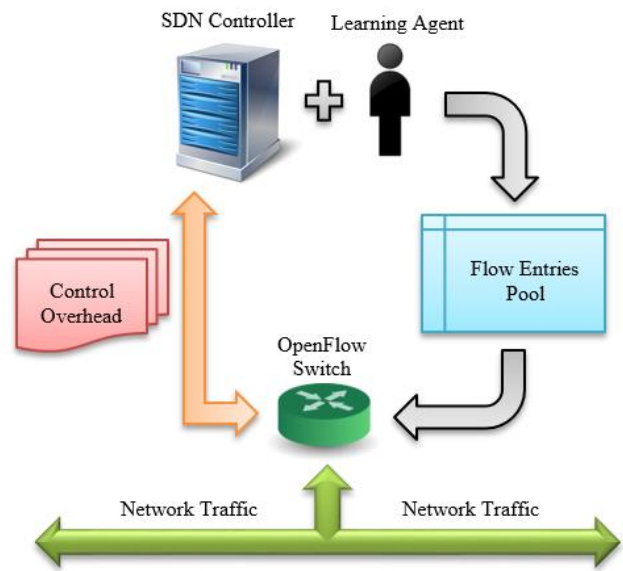

Fig. 1. Overview of our RL- based approach for managing SDN Flow Entry

Our RL agent deployed in the controller (depicted in Figure 1) utilizes two metrics to determine the entries that should be kept on the switch: the flow match frequency and the flow recentness (duration). These metrics are automatically recorded based on the OpenFlow specifications [3]. The goal of the agent is to maximize the long-term accumulated reward. The reward is taken to be inversely proportional to the configuration overhead and the number of table-miss events. Therefore, our RL agent splits the pool of flow entries into two parts: the local switch entries and the remote controller entries. This splitting is performed to reduce the control plane overhead given the TCAM size of the SDN switches. In addition, our proposed RL approach adapts to the monitored network traffic pattern.

To study the efficacy of our proposed RL-based scheme in minimizing the control overhead through appropriate parameter selection for rule placement in TCAM, we have conducted a systematic performance evaluation using Mininet [7], [8], a software-based network emulator that emulates a collection of network elements. Our experiments are based on two types of RL algorithms: namely, traditional RL and deep RL. Both types are used to drive the parameter tuning in our emulations to decide on the proper actions needed at each state in order to maximize the long-term accumulated rewards. Our emulation results show an improvement of around 60% in reducing the long-term control overhead, and around 14% increase in the table-hit ratio compared to the Multiple Bloom Filters (MBF) method [9] given a fixed size flow table of 4KB. *To the best of our knowledge, this is the first work that has utilized RL for managing flow entries in SDNs.*

The rest of the paper is organized as follows. In Section 2, we discuss the research background and related work. The design and implementation details are presented in Section 3; while experiments, analysis and evaluations are conducted in Section 4. Finally, the paper is concluded in Section 5.

## 2 BACKGROUND AND RELATED WORK

The research background and related works are discussed in this section. In particular, we provide a discussion on the OpenFlow protocol, OpenFlow switch characteristics and limitations, TCAM capacity utilization optimization methods, and the applications of RL in networking in subsections 2.1, 2.2, 2.3, and 2.4, respectively. We include a discussion on these topics since this background is essential for understanding our problem, which aims to utilize RL-based algorithms to split the forwarding rules between the OpenFlow switches and the controller in order to minimize the control traffic overhead in SDNs.

### 2.1 OpenFlow Protocol

OpenFlow is a communication protocol defined between an SDN controller and OpenFlow switches, programmatically allowing the direct access to the forwarding plane of network elements and the packets routing path through those network elements. Based on the OpenFlow Switch Specification provided by Open Networking Foundation (ONF) [3], one or more flow tables, and a group table are defined in an OpenFlow switch that can be used to perform packet forwarding.

In the OpenFlow protocol, the SDN controller can add, update, and delete flow entries either reactively (i.e., in response to the incoming packets) or proactively (i.e., in response to the flow entry timeout mechanism) [3]. There are multiple components (match fields, priority, counters, instructions, timeouts, cookies, and flags) in a flow entry used by the packet matching process. The match fields and priority components are combined to uniquely identify each flow entry in a specific flow table.

A packet matching pipeline process may occur if there are multiple flow tables defined in the switch. When an incoming packet is received by an OpenFlow switch, the packet is processed against each rule using the highest-priority match flow entry first, and then instructions are applied to update the match fields, the action set, and the metadata (information carried from one table to another) of the packet, and finally by sending the matching data and action set to the next table if needed. Additionally, a table-miss event is triggered whenever a packet does not match any of the flow entries in the flow table. A table-miss action is defined based on the table configuration on how to process unmatched packets. This includes operations such as forwarding the packet to the controller, dropping the packet, or passing it to another table. Table-miss flow entries can be added in the flow table for performing future proper table-miss actions. If a flow table does not have the table-miss flow entry, the unmatched packets are dropped [3].

### 2.2 OpenFlow Switch Characteristics and Limitations

In order to provide quality network and infrastructure services to individual users and cloud applications, a large number of network policies are typically deployed in DCNs. These policies are enforced through a collection of rules such as network security and traffic engineering for Quality-of-Service (QoS). Given the capabilities provided by SDN, the number of rules can easily reach hundreds or thousands depending on the scale of the datacenter. Therefore, it is crucial to devise an efficient way to store rules since the switch hardware is equipped with limited memory capacity.

The memory module of SDN switches is typically embedded in an Application Specific Integrated Circuit (ASIC) that performs hardware-based packet forwarding at the line-rate. The memory module that is typically used by most commodity SDN switches is called Ternary Content-Addressable Memory or TCAM [10]. This extension of the content-addressable memory (CAM) is different from random access memory (RAM) in that it requires a data memory address to be specified for accessing data. In CAM, a query is presented to the entire memory in one clock cycle for the data content itself and CAM returns a list of addresses where that data is stored. Based on this data retrieval mechanism and the parallel nature of CAM, its performance is considerably superior to the RAM data access process. Besides, the term "ternary" in TCAM refers to the ability of this memory module to utilize three different inputs (i.e., 0, 1, and X). This capability allows for broader search patterns in the entire memory. This search pattern process works well for networking applications, as it is similar to how net-masks operate. Compared to the binary CAM module that only performs exact pattern searches using only 0s and 1s, TCAM is commonly used in packet switching devices for additionally performing wildcard matching.

However, TCAM is expensive and consumes more energy compared to RAM (around $350 for 1 Mbits chip [11]). Currently, the cost of a TCAM module is around 400 times more than a RAM module with the same capacity, and its energy consumption is around 100 times that of RAM [11]. Due to the current technological limitations, the capacity of TCAM is usually around 1 ~ 2 Mbits [12]. The TCAM capacity for commercial grade networking equipment is typically around 18 Mbits [11]. According to the OpenFlow specification 1.4 [3], a flow entry consists of 40 tuples of which 13 are necessary tuples and 27 are optional. Generally, a flow entry occupies 356 bits [12] of memory if it contains 15 tuples. A TCAM module of 2 Mbits size can hold around 6,200 flow entries, while the average number of flows arriving at a switch is estimated to be 75,000 to 1.3 million flows per minutes, depending on the size of the network.

With this in mind, an SDN controller can easily exceed the TCAM limit of the managed switches by adding a large number of flow table entries. This is especially the case when other rules need to be maintained in the switch, including firewalling rules and Access Control Lists (ACLs) for security measures, or rules defining routing mechanisms and traffic monitoring [18]. Once the TCAM limit is reached, the OpenFlow switch stops accepting messages from the controller, which causes significant time delay for delivering packets since all incoming packets need to be processed by the controller if the packets cannot be matched locally by the switch. This induces an increase in controller-switch communications causing unnecessary traffic on the control plane of the network as well as significant delay in packet processing.

## 2.3 TCAM Capacity Utilization Optimization Methods

Since the number of forwarding rules increases proportional to the number of nodes attached to the network, the current limited TCAM capacity cannot accommodate all entries for large-scale networks with lots of nodes. This has been shown to be a proven problem for deploying OpenFlow enabled switches in production environment [18]. In order to mitigate this issue, several studies have been introduced in the recent literature. These studies can be organized under the following categories:

2.3.1 *Flow Table Compression*: The flow table compression technique is utilized by [12], [13], and [14]. The authors in [13] proposed to use the zero compaction technique to reduce the routing table size for IPv6 addresses in the following two steps. Firstly, the number of zeros in IPv6 addresses are replaced with a *don't care* statement (x) in the IP lookup table so that the size of the IP address can be further reduced. Secondly, prefix overlapping for the IP addresses having the same next hop IP addresses is applied, and only one of those overlapping prefixes that have the same next hop is stored. Zero compaction along with prefix overlapping and prefix minimization techniques in [13] can reduce the routing table size by 50% - 60%. In [14], the authors proposed a technique called "Multidimensional Table Compression" (MDTC) that uses the concept of a box to represent a flow entry in the table, and a set of boxes consist of a multidimensional table. This multidimensional space is then iteratively divided to obtain a set of small boxes, which are merged if they have the same action. The basic idea of this approach is to use a bigger box to cover multiple smaller boxes conducting the same action. Based on the proposed method in [14], the MDTC can reduce entries of ACLs by 23%, and the generated OpenFlow flow tables by 2.4%. The authors in [12] proposed a two level tagging approach that replaces flow entries with two layers of simpler and smaller tags in order to reduce the size of flow entries. The Tag-In-Tag consists of the PATH TAG (PT) that associates a tag with a given path used to route the packet; the FLOW TAG (FT) used to associate packets with a flow. Both PT and FT are used jointly so that the flows can be uniquely identified without using the original flow entry tuples. Based on [12], the proposed approach can store 15x more entries in a fixed size TCAM.

2.3.2 *Flow Entries Aggregation*: In [15], the authors proposed techniques to merge multiple flow entries so that the flow table usage can be minimized. The authors designed an offline scheme called Fast Flow Table Aggregation (FFTA), to shrink the flow table size, and an online scheme, called improved-FFTA (iFFTA), to update the flow entries efficiently. In the design of FFTA, this technique utilizes an idea similar to that used in the 3-step aggregation framework of bit weaving. This framework first divides the forwarding rule list into the Binary Search Tree (BST-based) prefix-permutable partitions, and then performs the Optimal Routing Table Constructor (ORTC-based) aggregation, and finally applies the bit merging process that iteratively merges rules together that differ by a single bit or have the same action into one entry on each partition to reduce the number of entries that reside in the flow table. Since the original flow table is divided into partitions and each partition is aggregated, the iFFTA is designed to perform flow entry updates efficiently by first updating the affected partition in the modified-BST, and then re-runing a bit merging process for the associated rules.

2.3.3 *Flow Table Management*: According to the design of the OpenFlow protocol, the flow entries can be removed from the flow table proactively or reactively. In a proactive mode, the entries can be moved by using a flow entry expiry mechanism, which is controlled by two attributes: *idle_timeout* and *hard_timeout* [16]. The *idle_timeout* enforces the flow entry to be removed when the flow entry is not used for a given time frame. The *hard_timeout* defines when the flow entry to be evicted from the flow table regardless of its status. In a reactive mode, the flow entries can be evicted by the switch itself. The work in [9] and [17] are built on top of this default flow expiry mechanism, and the authors in [9] proposed an autonomous and intelligent flow eviction mechanism to purge the flows using smart data logging with Multiple Bloom Filters (MBFs). MBFs are implemented in column-major order to perform logical shift operations on the importance value of flows, which is calculated based on the reference locality and recentness. Higher importance values lead to higher weights with the objective of reducing the switch-controller communications by automatically evicting less important flow entries from the switch. In [17], the authors proposed the TF-IdleTimeout approach to dynamically adjust the flow entry *idle_timeout* based on real-time network traffic. Their approach is based on analyzing the packet arrival intervals of real-time traffic and then adjusting the flow entry *idle_timeout* in order to effectively improve the utilization of the TCAM capacity.

2.3.4 *Flow Cache Mechanism*: In [11], the authors proposed a flow cache architecture that integrates a collection of hardware and software switches to work as if they were a single switch with unlimited forwarding rule capacity. This architecture uses one CacheMaster module to receive OpenFlow commands from the controller and distribute the forwarding rules to underlying switches. The forwarding rule placement algorithm in [11] uses the priority of the rule to order the collection of rules where each rule is associated with a match, action, and a weight value used to calculate the traffic volume matching the rule. The algorithm compiles a list of prioritized n rules with the goal of maximizing the sum of the weights of the rules installed in the TCAM. In [6], the authors proposed a cache architecture that utilizes a separate cache memory that is shared among multiple switches. This shared cache memory is placed between the controller and the underlying switches. Therefore, when required, a switch consults the flow cache memory first instead of communicating with the controller directly. To ensure fairness between mice and elephant flows, and to prevent the elephant flows from early eviction, the authors utilized the SHA hashing algorithm to ensure that the flows are mapped to the designated shared cache of each switch. For the cache flow replacement algorithm, the authors employed a localized least recently used (LRU) policy to evict flows from the cache memory where a mice flow can only evict a mice flow and an elephant flow can only evict an elephant flow.

Table 1. Approaches that Improve the TCAM Capacity Utilization Efficiency

| Approach Category | Summary |
| --- | --- |
| Flow Table Compression | <ul><li>The authors of [12] proposed the two-level tagging approach that replaces flow entries with two layers of simpler and smaller tags in order to reduce the size of the flow entries.</li><li>Work of [13] proposed to use zero compaction technique to reduce the routing table size for IPv6 addresses.</li><li>Authors of [14] proposed a "Multidimensional Table Compression" technique.</li></ul> |
| Flow Entries Aggregation | <ul><li>In [15], the authors designed an offline scheme, namely Fast Flow Table Aggregation (FFTA), to shrink the flow table size, and an online scheme, called improved-FFTA (iFFTA), to update the flow entries efficiently.</li></ul> |
| Flow Table Management | <ul><li>The authors of [9] proposed an autonomous and intelligent flow eviction mechanism to purge the flows using smart data logging with Multiple Bloom Filters (MBF).</li><li>In [17], the authors proposed the TF-IdleTimeout scheme to dynamically adjust the flow entry *idle_timeout* based on real-time network traffic.</li></ul> |
| Flow Cache Mechanism | <ul><li>The work of [11] proposed a differential flow cache framework that achieves fairness and efficient cache utilization with fast lookup.</li><li>The authors of [6] proposed a hardware-software architecture called "CacheFlow" with infinite rule capacity.</li></ul> |

Although SDN-enabled DCNs can dynamically manage the utilization of networking resources, it is difficult to determine which forwarding rules should remain in the switch memory and which should be processed by the controller with the objective being to minimize information exchanged between the controller and the switches. Moreover, it is hard to prevent the overuse of the TCAM capacity by the controller without constantly monitoring it, which implies that more information exchange overhead is needed by this proactive process. Alternatively, our proposed approach can address these issues using machine learning techniques to automatically determine which forwarding rules should remain inside the switch memory so that the overall network control overhead is reduced regardless of the traffic pattern within the DCN, without proactive sampling, logging, or monitoring of the incoming packets.

## 2.4 Applications of RL in Networking

The rise of machine learning techniques has motivated networking researchers to apply these techniques in networks. In particular, RL—a machine learning paradigm in which a learning agent learns what actions it should choose while interacting with a dynamic system so that cumulative reward is maximized—is emerging as a popular machine learning technique with numerous applications in networking [19], [20] (e.g., routing [29]). In RL, an iterative learning process is conducted by making an agent who receives the current state and the reward from the dynamic system, and then performs the associated action based on its gained knowledge in order to develop a policy to maximize the obtained long-term cumulative rewards via state transitions.

In the work of [30], the authors proposed a new machine learning approach to choose the optimal route in ad hoc networks, which is based on the cooperative RL in modeling the swarm intelligence from the models of social insect behavior. To prove the effectiveness of their proposed approach, the authors presented the analysis and performance evaluation by comparing with the existing routing protocols, and the results indicate that the packet delivery ratio is significantly improved when RL is utilized.

In [31], the authors presented some extensions based on the Ad hoc On-Demand Distance Vector (AODV) routing protocol, which emphasized on improving the route error tolerance mechanism. The proposed technique was based on the fact that a link failure in the middle of the transmission when the transmission started from the closest neighbor node. The simulation results of this approach demonstrate the successful reduction in transmission delay and improvement of the packet delivery ratio in comparison with the traditional AODV. The results also show the improvement in reducing the route discovery frequency to minimize the routing overhead.

In the SDN domain, RL was proposed for improving the efficiency of the network routing in [32], where authors proposed a QoS-aware adaptive routing (QAR) in the three-level multi-layer hierarchical large SDNs including the super, domain, and slave controllers. This design shares the control loads among the controllers and can reduce signaling delay significantly. With the aid of a RL-based algorithm, their simulation results show that QAR performs better compared to existing learning solution and provides faster convergence rate in QoS provisioning.

In addition, the work of [33] implemented a deep RL algorithm, which is similar to our proposed technique, a Deep Q-Network (DQN) to deal with large state space in terms of several correlated channels in a Wireless Sensor Network (WSN). The objective of this design is to maximize the expected long-term reward in obtaining the near optimal channel used for packet transmission without failing, and the simulation results indicate that DQN can achieve near-optimal performance in dealing with complex real scenarios that do not necessarily present Markovian dynamics.

There are several proposals where RL is deployed over the existing routing protocols. However, none of them use RL in flow entry management in SDNs. As far as we know, we are the first work that utilizes RL algorithms, both traditional RL and DQN, in managing flow entries in SDNs. Our approach utilizes RL, along with a fusion of traditional RL and neural networks, namely DQN, which has the capability of handling more complex systems and results in better performance [21], [22], [23]. Table I summarizes previous approaches and existing research work that addresses the efficiency for TCAM capacity utilization for OpenFlow enabled switches.

Among the aforementioned related works, our proposed approach is similar to the work of [9] since both approaches utilize frequency and recentness in their algorithms with the goal of reducing the network control overhead, while increasing the table-hit ratio. In Section 4, a performance comparison study between our proposed RL-based approach and the work in [9], the Multiple Bloom Filters (MBF-based) approach is conducted.

## 3 DESIGN AND IMPLEMENTATION

In order to effectively emulate an SDN-enabled networking environment, we use Mininet [7], [8], a software-based network emulator that emulates a collection of end-hosts, switches, routers, and links on a single Linux kernel, allowing the entire OpenFlow network to be emulated on a single computer with lightweight process-based virtualization. In the following subsections, we first introduce the learning framework, then discuss the design of our approach, and finally provide the details of our proposed algorithms.

### 3.1 Design of the Learning Framework

In our approach, the configuration of rules can be defined by two decisive parameters, *flow match frequency* and *flow recentness*. Therefore, the RL algorithm can be modeled to obtain the network configuration that minimizes the long-term control plane overhead. This approach is modeled using a *Markov Decision Process* (MDP) [19], which can be represented as the tuple <S, A, P, R>, where $S = \{s_1, s_2, ..., s_i\}$ is the state-space, $A = \{a_1, a_2, ..., a_j\}$ is the action-space, *P* defines the transition probability from state *i* to state *j*, and *R* defines the reward associated with various actions $a \in A$. The goal of this MDP is to develop a policy $\pi : S \to A$ that maximizes the cumulative rewards obtained in the long-term. In the problem that we address in this research, we can know all the forwarding rules that need to be processed for all traffic flows transported on the network. For our problem, the state-space, action-space, and reward function are defined as follows:

- **State-Space**: The state-space in our model represents all possible combinations of values of the flow match frequency and the flow recentness parameters. We denote the state of the chosen set of parameters as follows:
$$s_i = (flow\_freq_i, flow\_recentness_i)$$
where $flow\_freq_i$ represents the frequency that a given flow is matched, whereas $flow\_recentness_i$ indicates the residence duration of the given rule in the switch's memory.
- **Action-Space**: The action-space in our approach includes the following options: (1) no action; (2) increase value associated with the chosen parameter; and (3) decrease value associated with the chosen parameter. For example, we denote the increase action on the flow frequency parameter as follows:
$$a_{flow\_freq_i}^{increase} = (flow\_freq_i^{increase}, flow\_recentness_i)$$
- **Reward Function**: The network configuration overhead is defined as the number of communications induced between the controller and the switches in order to properly forward the input packets. The reward is determined based on the measured configuration overhead using the given selected forwarding rules installed in the switch, which is represented as follow:
$$r_t = Compare(overhead_{current\ best}, overhead_t)$$
where $overhead_{current\_best}$ denotes the current best network control overhead obtained so far by the emulation. There are three types of rewards (i.e., 1, 0, and -1). For a given comparison, a configuration with less overhead returns a positive reward 1 to the learning agent; otherwise, the agent is given a negative value -1 as the reward. If $overhead_t$ and $overhead_{current\_best}$ are equal then a 0 reward is given.

In our approach, we utilize the following RL algorithms:

3.1.1 *Traditional RL*: In our design, the *temporal difference* (TD) Q-Learning algorithm is utilized in our emulation experiments. The decision to utilize the Q-Learning algorithm is motivated by the fact that it does not need a model for the environment and it updates the Q-values at each time step based on the estimation of action-value function. This iterative learning process is conducted in discrete time steps in which the agent interacts with the environment. At each step, the agent selects an action $a_t \in A$ in the given state $s_t$. The agent makes its own decisions to choose an action based on the action selection policy with the goal of maximizing the expected reward. The notation $Q(s, a)$ denotes the average Q-value of an action *a* at state *s*. While the immediate reward is collected, $Q(s, a)$ can be further refined as:

$$Q(s_t, a_t) = Q(s_t, a_t) + \alpha * [r_{t+1} + \gamma * \max_a Q(s_{t+1}, a_{t+1}) - Q(s_t, a_t)] \tag{1}$$

where $\alpha$ ($0 < \alpha \leq 1$) is the learning rate that determines to what degree the newly obtained information overrides the old one, and $\gamma$ ($0 \leq \gamma \leq 1$) is the discount factor that defines the importance of future rewards and guarantees convergence of the accumulated reward. In our experiments, we use a learning rate of 0.1 and a discount factor of 0.95 as these values are commonly used in practice [19].

---

**ALGORITHM 1**: Q-Learning Algorithm
**Pre-condition**:
    Initialize Q table with small random number
    Initialize state $s_t$
    Initialize goal $\mu$
**Procedure**:
01: *improvement* = 0
02: **repeat**
03:   **for** (step = 0; step < learning_iteration; step++)
04:     Get action $a_t$ from $s_t$ using $\epsilon$-*greedy* policy
05:     Get parameter $param_t$ from $s_t$ using $\epsilon$-*greedy* policy
06:     $\epsilon = \epsilon - $ (step / learning_iteration) * $\epsilon$
07:     Take action $a_t$ on the $param_t$ and receive reward *r*, control overhead *c*
08:     Sample new state $s_{t+1}$ after applied action $a_t$
09:     Update $Q_t \leftarrow Q_t + \alpha * (r + \gamma * maxQ_{t+1} - Q_t)$
10:     Update the corresponding $param_t$ of $Q_t$
11:     *improvement* = get_improvement($best_t, worst_t$)
12:   **end for**
13: **until** improvement > $\mu$

---

The $\epsilon$−*greedy* policy is used as the action selection policy in which the action with the highest Q-value is always picked while the probability of picking some other action at random is small. This implies that the algorithm can have a chance to explore the action space in the process of searching for a better action that produces a better reward. Besides, the algorithm also keeps track of the chosen parameter corresponding to the Q-value of each action. If the action is chosen to be performed due to its Q-value, the associated parameter value will be updated as well. This Q-Learning algorithm is presented in Algorithm 1 where *μ* represents the terminal objective of the algorithm.

3.1.2 *Deep RL*: The deep Q-network (DQN) algorithm is a deep RL algorithm that utilizes a class of artificial neural networks (ANNs) known as deep neural networks [21]. ANNs are used to approximate functions by learning from a large set of input data and the term "*deep neural network*" (DNN) refers to a neural network with multiple hidden layers [24]. DQN uses feedforward neural networks with three basic components: (1) *Neurons* interconnected using directed links to form a network; (2) *Weights* associated with each connection; (3) *Layers* consisting of a number of neurons (including multiple hidden layers).

Neural networks are usually represented in the form of multi-layers of neurons in a directed graph with fully connected synapses between layers. The input data is fed through the *input layer* (the first layer of the network) and the output is produced at the output layer (the last layer of the network). The *hidden layers* between the input and output layers are responsible for intermediate calculations. The size of the input data and the complexity of the problem being solved determine the number of neurons needed in each layer. In practice, there are no fixed rules defining how many hidden layers or how many neurons should be used in each layer of the DNN and these decisions must be made according to the context of the problem. Deeper neural networks have more capacity to learn higher levels of abstractions from the training data. Furthermore, an activation function is implemented to compute the activation value and propagate it between connected neurons where each synapse between two neurons is associated with a numeric weight value. A weighted sum of all the inputs to neuron $j$ needs to be calculated in order to compute the activation value $a_j$ by applying an activation functions $f$ as follows:

$$a_j = f(\sum_{i=0}^{n} w_{i,j} * a_i) \qquad (2)$$

There are several variations of activation functions to choose from based on the application. These functions include *Sigmoid*, *SoftMax*, *Rectified Liner Unit* (ReLU), etc. ReLU is gaining popularity in DNNs due to its simplicity and its effectiveness in dealing with the vanishing gradient and sparsity issues [24]. The ReLU activation function is defined as follows:

$$f(x) = \begin{cases} 0, \text{for } x < 0 \\ x, \text{for } x \geq 0 \end{cases} \qquad (3)$$

In our emulation experiments, we assume that it is impractical to process all states since some of them are never or rarely chosen which leads to longer convergence times for the Q-table. Therefore, it is a good practice to approximate (predict) the Q-values for extreme states instead of processing them. This prediction process is based on previously processed states (i.e., during the training phase). Therefore, we can endow our original Q-function with a deep neural network that takes the state (i.e., flow match frequency, recentness, chosen action-parameter pair, and the obtained reward) as an input, and generates the Q-values for all possible actions and its corresponding chosen parameter. The DQN update process in the *i*th iteration is formulated by the loss function as follows:

$$L_i(\theta_i) = \mathbb{E}_{(s,a,r,s') \sim P(D)} \left[ \left( r + \gamma \max_{a'} Q(s', a'; \theta_{i-1}) - Q(s, a, ; \theta_i) \right)^2 \right] \qquad (4)$$

In equation (4), $\mathbb{E}_{(s,a,r,s')}$ is the processed experience which is uniformly sampled using probability $P(D)$ from the experience memory. The experience memory stores all experienced transitions $(s, a, r, s')$, where $s$ is the current state, $a$ is the action selected to perform, $r$ is the reward received, and $s'$ is the next state after the action is applied. In addition, in (4), $\gamma$ is the discount factor, and $\theta_i$ is the state of the Q-network in the *i*th iteration, whereas $\theta_{i-1}$ is the state used to calculate the target in the *i*th iteration. Similar to traditional RL, we use a discount factor of 0.95 in our experiments.

---

**ALGORITHM 2**: Deep Q-Network Algorithm

**Pre-condition**:
    Initialize expereience memory $M$
    Initialize action-value pair $Q$ with random weights
    Initialize state $s_t$
    Initialize goal $\mu$

**Procedure**:
01: *improvement* = 0
02: **repeat**
03:   **for** (step = 0; step < learning_iteration; step++)
04:     Get action $a_t$ from $s_t$ using $\epsilon$-*greedy* policy
05:     Get parameter $param_t$ from $s_t$ using $\epsilon$-*greedy* policy
06:     $\epsilon = \epsilon - (\text{step} / \text{learning\_iteration}) * \epsilon$
07:     Take action $a_t$ on $param_t$ and receive reward $r$, control overhead $c_t$
08:     Observe new state $s_{t+1}$
09:     Store expereienced memory $(s, a, r, s_{t+1})$ into $M$
10:     Sample $n$ random transitions $(s', a', r', s'')$ from $M$
11:     Update *transition* $tt \leftarrow r' + \gamma * \max(s'' - a'')$
12:     Update the $param_t$ of $\theta_i$
13:     Train the $Q$ network using $loss = \left(tt - Q(s', a')\right)^2$

---

| **ALGORITHM 2**: Deep Q-Network Algorithm |
|---|
| 14:   $improvement = \text{get\_improvement}(best_t, worst_t)$ |
| 15:   **end for** |
| 16: **until** $improvement > \mu$ |

The ϵ−greedy policy is also used as the action selection policy to select the action based on the highest Q-value associated with that action. The ϵ value is set to 1 at the beginning of the learning process so that the agent can choose random actions, and the value is then decreased over time in order to maintain a fixed exploration rate. Similar to the original RL, the algorithm will also keep track of the chosen parameter corresponding to the Q-value of each action. The Deep-Q-Network (DQN) algorithm is presented in Algorithm 2 where $\mu$ represents the terminal objective of the algorithm.

## 3.2 Problem Formulation

Searching for the best parameter set that minimizes the control overhead in our emulated topology is an instance of the *set partition problem*, which is known to be an NP-Hard problem. In this section, we formally present the problem addressed in this research as an Integer Linear Programming (ILP) problem. (However, since ILP problems are intractable for large-sized problems, we will instead use an RL-based solution, which works better in practice for larger networks.)

Let $T = \{t_1, t_2, \dots, t_N\}$ be a multiset where element $t_i$ is the rate of overhead traffic that is induced over the control plane of the network when the $i^{\text{th}}$ forwarding rule is installed on the SDN controller. Otherwise, the rate of flow traffic induced on the control plane is zero when the forwarding rule is installed on the switch. Moreover, let $S = \{s_1, s_2, \dots, s_N\}$ be a multiset where element $s_i$ is the size of the $i^{\text{th}}$ forwarding rule. Finally, let $a_i^c$ and $a_i^s$ be selectors with a value of 1 if the $i^{\text{th}}$ forwarding rule is assigned to the SDN switch or controller and 0 otherwise. Then, our problem can be formulated as follows:

$$Minimize \sum_{i=1}^{N} t_i \cdot a_i^c \tag{5}$$

Such that

$$\sum_{i=1}^{N} s_i \cdot a_i^s \leq C_{TCAM} \tag{6}$$

$$a_i^c + a_i^s = 1 \; \forall_i \; 1 \leq i \leq N \tag{7}$$

$$a_i^c \in \{0, 1\} \; \forall_i \; 1 \leq i \leq N \tag{8}$$

$$a_i^s \in \{0, 1\} \; \forall_i \; 1 \leq i \leq N \tag{9}$$

where $C_{TCAM}$ represents the capacity of the SDN switch TCAM and $N$ represents the number of forwarding rules that need to be installed on the SDN switch or controller. The objective function (5) strives to minimize the rate of overhead traffic induced on the control plane of the SDN network. Constraint (6) ensures that the total size of the forwarding rules installed on the switch does not exceed the capacity of the switch TCAM. Constraint (7) ensures that each forwarding rule is installed on the switch or the controller. Moreover, this constraint ensures that no forwarding rule is left uninstalled or duplicated. Finally, constraints (8) and (9) indicate that the decision variables can only take integer values of a 0 or 1.

## 3.3 Deep Learning Framework: Caffe

In our experiments, we employ the open source deep learning framework, Caffe [25], as our deep learning engine. It is developed and maintained by the *Berkeley Vision and Learning Center* (BVLC) using the C++ programming language. The Caffe framework provides a clean and extensible toolset that exposes the functionality of deep learning algorithms to its users, and it also utilizes the GPU technology to further optimize the computational efficiency and execution time. The framework supports both Python and MATLAB interfaces besides C++, and it has the following main components:

- **Blobs**: Caffe utilizes the 4-dimentional array, blob, as the unified memory interface for storing data such as images, model parameters, etc. An excellent feature the blobs provide is the masking of low-level memory operations in CPU or GPU by synchronizing the information between them as needed, without having to access low level Application Programming Interface (APIs) to manage memory addresses or space allocations.
- **Layers**: The layer model in Caffe maps to a neural network layer that takes one or more blobs as input, and produces one or more blobs as output. The two main functions that a Caffe layer provides are: (1) The *forward pass* that produces an output based on the fed input; (2) the backward pass that takes the gradients of the output to compute the parameters of the inputs and propagates them back using the backpropagation process. There are several types of layer models supported by Caffe, including convolutional, pooling, and so on, with more details available in [25].
- **Network**: The Caffe network is the abstraction of the representation of a set of connected layers. It has built-in validation mechanisms to enforce the correctness of the constructed network. The network starts by loading data into the layer component and produces the results in the loss layer that can be further used in classification or pattern extraction.

Our deep RL algorithm implementation is based on the work of Mnih et al. [21], [22] with few customizations tailoring it toward our research focus. The overall architecture of our DQN approach, as depicted in Figure 2, consists of one input layer that takes the state space as input, three fully connected inner product layers (hidden layers), and the loss layer that generates the output. The output in our approach maps each configuration state to one of three actions: no action, increase action, and decrease action with its corresponding parameter (i.e., flow match frequency or flow recentness).

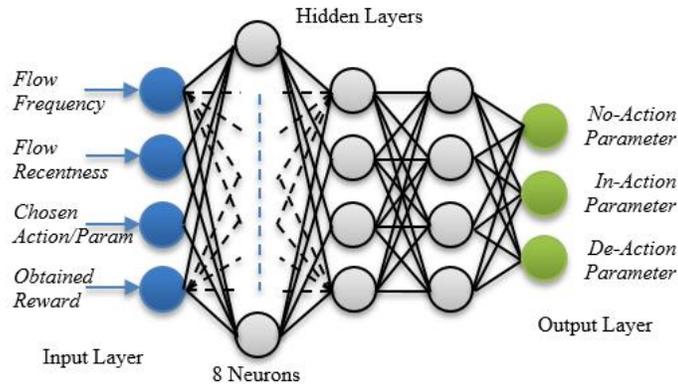

Fig. 2. Deep RL Neural Network Structure

## 3.4 Traffic Flow Generation

The traffic used in our emulation experiments is based on flows. Flows can be further classified into two categories: elephant flows that contain high volume data and mice flows that carry small amounts of data. The flow type generated has the ratio 1:9, where 10% of the flows are elephants and 90% are mice. According to [26], the size of a mice flow is on average 256 KB, and the size of an elephant flow is considered as tens of megabytes. In our experiments, 25.6 MB and 256 KB were utilized for the elephant and mice flows, respectively, and these flows are transmitted over the 1 Gbps links using TCP/IP.

Mininet provides functionality to execute Python scripts within each emulated node. Therefore, each node can run the script and send TCP/IP traffic to a designated destination on a predefined port number. The traffic destination generated by each node is determined based on the passed-in arguments for the purpose of the flexibility. Mininet's built-in "*iperf*" tool is also employed jointly with our script in order to create random traffic patterns to simulate the DCN traffic. Our traffic model assumes a greedy source generating Poisson traffic at an average rate of 310 Mbits/sec.

## 3.5 Mininet Emulation Design

For our emulation experiments, we designed a simplified model that utilizes one SDN controller and one OpenFlow switch connected with 20 hosts. This simplified topology can generate more than 4000 forwarding rules to be stored in the OpenFlow switch, which by far exceeds the capacity of a 1 Mbits TCAM module. Some of the 20 emulated nodes send traffic while the others act as receiving hosts as depicted in Figure 3. The configuration overhead is based on counting the number of times the controller communicates with a switch in order to forward a packet to its destination.

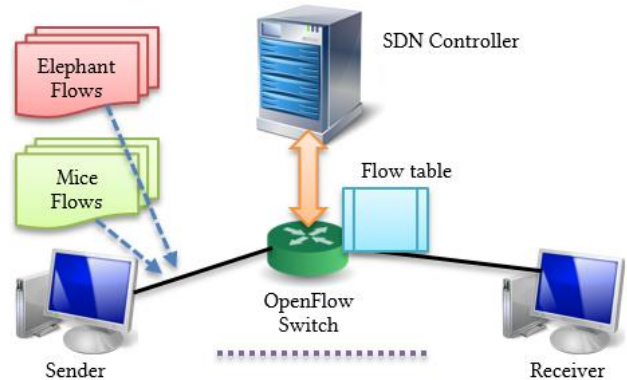

Fig. 3. Mininet Emulation Topology

In our emulation experiments, we utilize POX [27], a Python based networking software platform, as our SDN controller. POX is an open source general SDN controller that supports OpenFlow protocol version 1.0, and provides rich APIs that expose the core functionalities of POX. We implemented a customized POX component and extended the following POX built-in event handlers:

- *ConnectionUp*: This event is triggered in response to the establishment of a new control channel with a switch. We extend this event handler by adding the code to install the selected forwarding rules created by the learning agent prior to handling any incoming traffic.
- *ConnectionDown*: Similar to the *ConnectionUp* event, this event is fired when a connection to a switch is terminated, either explicitly or caused by switch rebooting. We utilize this event to handler to output the final value of measured control overhead of the emulation session. This value can be used later to determine the reward of the learning agent for the session.
- *PacketIn*: This event is raised whenever the controller receives an OpenFlow packet-in message from the switch. This indicates that a packet has arrived at the switch port and the switch either does not know how to handle this packet (i.e., it failed to match all forwarding rules in its tables), or the matching rule action implies sending the packet to the controller. This is where the controller performs the overhead calculation by incrementing the overhead when the switch forwards a packet.

At the beginning of each emulation experiment, traffic flows are generated into the network topology and the state of the flow table in the switch is observed. To reduce the learning overhead, the deployment of our proposed approach is configured into two phases: orchestration phase and production phase. The orchestration phase is where the state-space data collection process is initiated to gather all the forwarding rules needed to properly process all the incoming traffic. During this data collection process, the flow pool is constructed on the controller to record all the unique flow entries for a given time interval after the traffic has been injected into the network. We are aware that the orchestration phase has significant influence on the performance of the production phase because the network traffic patterns vary with time and with the applications running on top of the network. The implementation of a data collection process that adapts to different network traffic patterns during the orchestration phase will be addressed by our ongoing research. The duration of the data collection phase of our experiments is 5 minutes.

During the production phase, the RL agent is deployed on the controller to automatically search for the optimal network forwarding rules from the flow pool based on two criteria: flow match frequency and recentness. Therefore, the agent determines rule configurations that minimize the control plane overhead. The agent produces a configuration that fits the traffic patterns monitored on the network.

As depicted in Figure 4, the workflow of our proposed approach is presented as follows:

1) Based on the decision made by the learning agent, the parameters *flow_freq* and *flow_recentness*, are used as the forwarding rules selection criteria from the pool of all rules.
2) The chosen forwarding rules are installed in the flow table of the switch before it starts to handle the traffic.
3) The emulation process is initiated and control plane overhead is reported at the end of the emulation session.
4) The agent computes the reward for the current forwarding rule configuration and updates the Q-values and its corresponding parameters.
5) The above steps are repeated until the goal is satisfied.

Once the learning agent has found the near optimal parameters for selecting the forwarding rules that need to be placed on the switch, the overall network control plane overhead can be minimized while taking the capacity of the TCAM module into consideration.

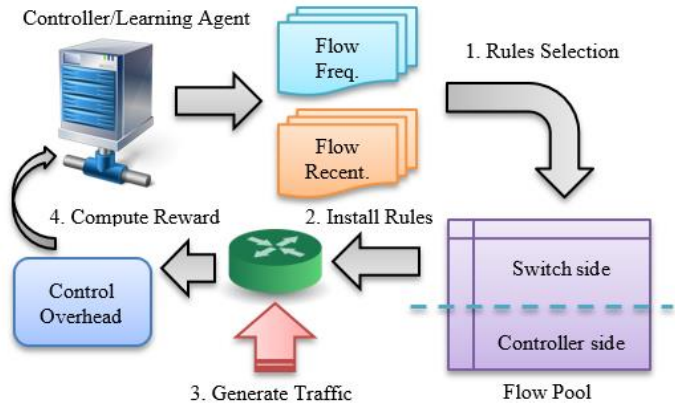

Fig. 4. Workflow of Proposed Approach

## 4 EXPERIMENT ANALYSIS AND PERFORMANCE EVALUATION

To demonstrate the utility and performance of our proposed RL-based approach, we evaluate our algorithms by conducting systematic experiments to compare the traditional RL and deep RL methods against a baseline that utilizes the Multiple Bloom Filter (MBF) method presented in the recent literature [9]. The performance metric considered in our study is the control plane overhead of each emulated experiment. Our proposed RL-based learning algorithms are deployed on the SDN controller as an added layer of intelligence.

To evaluate the effectiveness of our RL-based approach, we utilize two Mininet virtual machines (VMs) running on top of a *VirtualBox* hypervisor. The VMs implement the traditional RL and DQN agents. The entire framework that we developed for our experiments is compiled with Python version 2.7 on Ubuntu Linux LTS 14.04. The Mininet version used in our study is 2.2.1.

### 4.1 Performance of Traditional RL

In this section, we discuss the effectiveness of our proposed RL-based approach that strives to identify a near optimal configuration that minimizes the overall control plane overhead while increasing the switch memory usage efficiency.

For our first experiment, we studied the performance of the traditional RL on two different sets of decision parameters in a network with a light traffic volume. The first set (Figure 5a) has values of 90 and 30 for the flow match frequency and the flow recentness parameters, respectively. This implies that any forwarding rule with a match frequency more than or equal to 90 or recentness less than or equal to 30 seconds will be chosen to be placed into the flow table on the switch. Another set (Figure 5b) of parameters has values of 40 and 50 for the flow match frequency and the flow recentness parameters, respectively. The objective of this experiment is to evaluate the adaptability of the traditional RL agent to different sets of parameters. As presented in Figure 5, the emulation results show the performance improvement in terms of control plane overhead for those two different sets of parameters.

In this experiment, the goal of the traditional RL agent is to obtain the parameter set that can be used to choose the forwarding rules that have at least 40% of reduction in terms of network control plane overhead. The results show that the control plane overheads of these two different sets have been reduced down to at least 40% as the agents learn over time, and the network control plane overhead has been minimized more than 40% compared to the initial control plane overhead. The 40% goal is chosen in this experiment but other goals can be chosen as well. The agent will run indefinitely if it is provided with an infeasible goal as it will continue to strive to achieve the given goal.

It is worth mentioning that in some cases the control overhead obtained by the RL agent decreases towards the end of learning episodes. For instance, the control plane overhead starts to deteriorate after $300^{th}$ episode for the agent shown in Figure 5 (b). We have determined that this is due to the nature of the $\epsilon$–$greedy$ policy that has a probability $\epsilon$ to pick some other action at random, regardless of its Q-value. Despite the fact that the exploration rate $\epsilon$ decreases as number of learning episodes goes up, the learning agent still has the possibility to perform an action that deviates from the known best action.

4.1.1 *The Effectiveness of Initialized Q-table*: The number of configurable parameters and the value range of each parameter define the size of the state-space of the RL algorithm. This size grows exponentially as the number of parameters and value range increase. Based on the characteristic of the RL algorithm, the learning agent performs a large number of explorations before it finds the goal state. We argue that this significantly degrades the efficiency of the RL algorithm.

In our second experiment, we again randomly picked two different sets of parameters that are not part of the Q-table initialization sets, with the agent given a goal to find the configuration that results in at least 40% reduction in the control plane overhead. In this experiment, we also observe the effect of the initialized Q-table (i.e., initialized policy) on improving the performance of our proposed approach. The motivation behind this experiment is that networks typically experience similar traffic patterns repeatedly. Therefore, we argue that our RL-based approach can achieve the goal faster and minimize the time and computational resources required to the required goal.

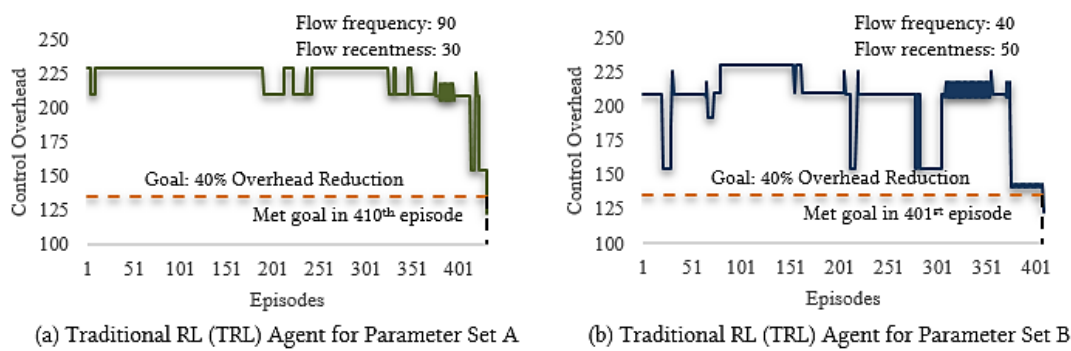

Fig. 5. Traditional RL agent with different parameters.

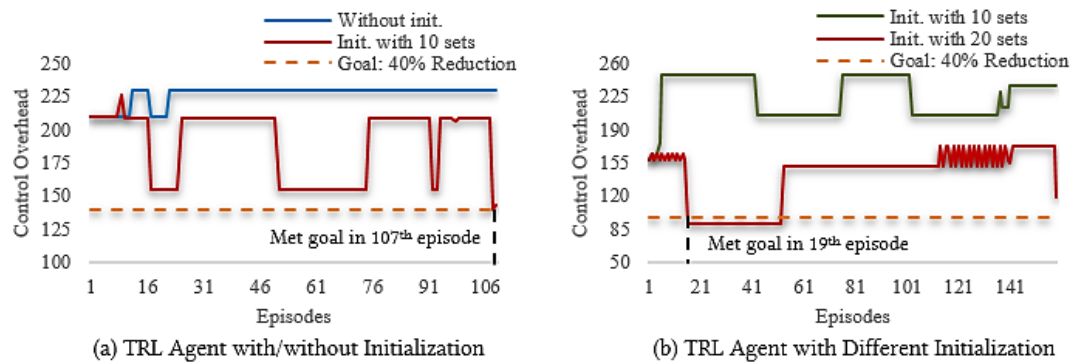

Fig. 6. Traditional RL Agent with and without Initialized Q-Table.

Figure 6 presents the performance of the traditional RL-based agent with and without initialized Q-table. As depicted in the figure, we compare the difference under the condition that the Q-table has been initialized with the training sets or no initialization at all. Figure 6 (a) shows how the initialized policy can lead the agent to obtain at least 31% control plane overhead reduction on the 17th learning episode. The agent also finds a configuration with around 40% reduction in the control plane overhead on the 106th episode. On the other hand, the agent without an initialized Q-table could not find the set of parameters that satisfy the goal within a budget of 150 learning episodes; thus, resulting in a longer convergence time.

In this experiment, we further analyzed the impact of the quality of the initialized Q-table on finding the optimal network control plane overhead in terms of learning episodes. Figure 6 (b) shows the agent's results when the Q-table is initialized with 10 and 20 different sets of parameters, respectively. We can learn that the agent with the Q-table initialized with 20 different sets yields a configuration set that achieves around 44% control plane overhead reduction on the 18th episode. This demonstrates that Q-table initialization using larger training sets can significantly improve the convergence time to a goal state that achieves the target improvement.

During the evaluation of the initialized Q-table in this experiment, we used a static exploration rate as small as 0.1 to let the learning agent perform less exploration, and allowing it to exploit the knowledge encapsulated in the Q-table.

## 4.2 Performance of Deep RL (DQN)

In this section, we present results related to the experiments that we conducted to study the performance of deep RL in support of finding a configuration set that minimizes the network control plane overhead. Our emulation experiments utilize the Caffe library to build a deep-Q-network. Similar to our traditional RL-based approach, the DQN-based RL agent approach explores the sets of parameters based on what it has learned during the training process.

Instead of evaluating all the actions individually from the previous transitioned state as implemented in the traditional RL agent, the DQN is utilized to approximate the possible Q-values and its corresponding parameters for all the actions for the given configuration set obtained from the previous state transitions.

Before the DQN-based RL agent interacts with the environment through emulation, the underlying neural network is constructed with 4 neurons in the input layer (two decision parameters, the chosen action, and the immediate reward), and 3 neurons in the output layer corresponding to each action with associated Q-value and parameter. The number of hidden layers is the free parameter that can be determined based on the user's preference. In our study, we consider 3 hidden layers between the input and output layers (Figure 2).

During the training phase, the *Stochastic Gradient Descent* (SGD) method is utilized to calculate the loss value. A vector of memory is used in the emulation to collect all the processed states associated with the action performed and the chosen parameter. These process states are referred to as the state transitions in our approach. The algorithm begins to update the Q-network by randomly selecting 4 transitions when the number of transitions inside the memory reaches a certain threshold, and uses those chosen transitions to calculate the weights in the network. The Q-network update process will continue until the agent reaches the goal state. In this experiment, the goal state is defined as the parameters set that produces at least 40% of control overhead reduction compared to the initial state.

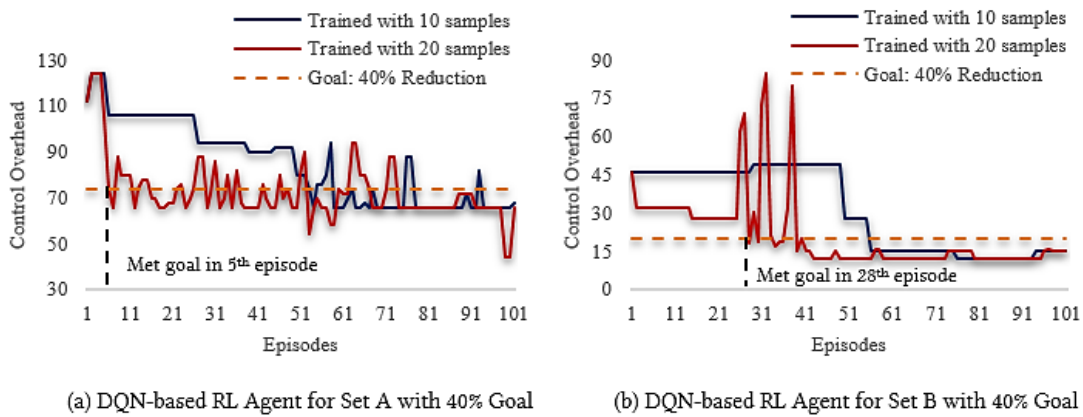

Fig. 7. DQN-based RL agent with different training scenarios.

To evaluate the performance of our proposed DQN approach, we designed two training scenarios and observed the performance of the DQN agent trained with 10 and 20 uniformly selected random sets of parameters. In our experiments, we used the same light traffic pattern that we used to evaluate the performance of the traditional RL agent with two different parameter sets.

By observing the performance of the DQN-based RL agent depicted in Figure 7, we can see that the learning agent trained with 20 randomly chosen samples outperforms the one trained with 10 samples. In this experiment, the goal state was reached within 5 learning episodes whereas another scenario required 28 or more episodes to research the goal state as shown in Figure 7 (b). This demonstrates that the DQN agent's convergence time improves with more training samples.

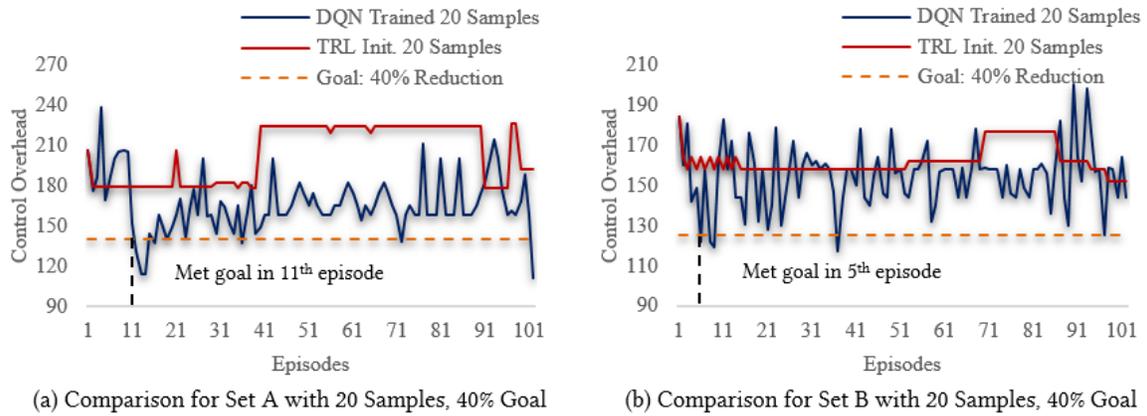

Fig. 8. Comparison between learning agents trained with 20 samples and a goal of 40% control plane overhead reduction.

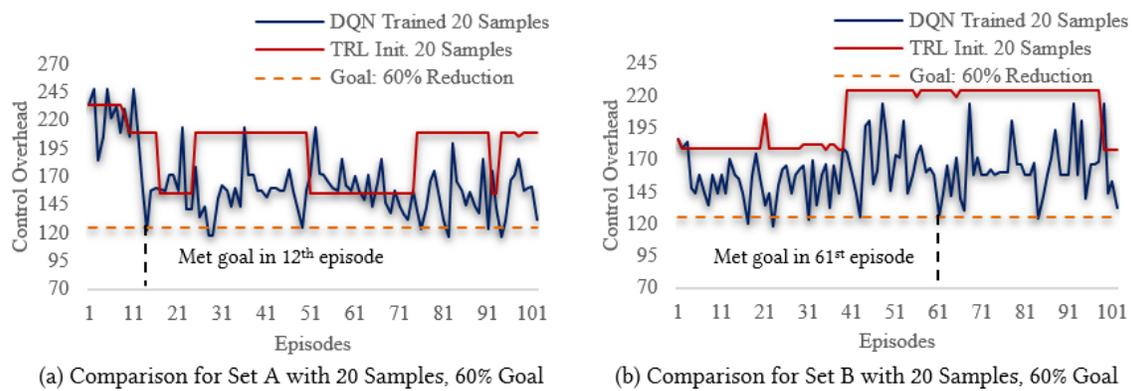

Fig. 9. Comparison between learning agents trained with 20 samples and a goal of 60% control plane overhead reduction.

During the evaluation of the initialized Q-table in this experiment, we used a static exploration rate as small as 0.1 to let the learning agent perform less exploration and more exploitation of its learned knowledge (i.e., using that Q-network). This setting is similar to the one we used in our traditional RL experiments.

## 4.3 Performance Comparison between the RL Agents

In this section, we present the results of the experiments we conducted to compare the performance of the traditional RL and DQN agents. To perform a fair comparison between the traditional RL and DQN agents, we designed a training/initialization scenario with 20 uniformly random parameter sets. The sets used to evaluate the agents do not belong to the sets that are chosen to train them. In contrast to our previous experiments, we use one light traffic pattern to train/initialize the Q-matrix, and use another different heavier traffic pattern to conduct the performance evaluation. We also experiment with two different objectives, 40% and 60%, to assess the control plane overhead reduction capabilities of the two agents.

In our experiments, we are interested in how the experienced agents react to new unseen traffic patterns and initial parameter sets that are not part of the training sets. The agents' evolution from this initial state completely relies on their experiences and the knowledge obtained during the training phase. Four randomly selected parameter sets are used as the initial state for the learning agents to start the emulation with the goal states being to achieve 40% and 60% reduction on the control plane overhead, respectively.

As illustrated in Figure 8, it is observed that the performance of the DQN-based agent is always better than the traditional RL agent in our experiments when the goal is set to achieve 40% control plane overhead reduction. From Figure 8 (a) and (b), it can also be noticed that the DQN-based agent is capable of finding a better set of parameters in a shorter period of time in terms (i.e., fewer episodes are needed to converge to the goal state). Moreover, as depicted in Figures 9 (a) and (b), the DQN-based agent continued to outperform the traditional RL agent when the goal is set to achieve 60% control plane overhead reduction. Thus, our experiments show that DQN-based agent outperforms the traditional RL agent when it is trained based on the same parameter sets, initialized similarly and has the same predefined objective.

From the results presented in Figure 8 and Figure 9, it can also be seen that a traditional RL agent could not find the parameters set that satisfies the predefined objective within the limited number of episodes (i.e., 100 episodes), whereas the DQN-based agent can satisfy the goal in fewer episodes. It is worth mentioning that the agent can stop as long as it satisfies the objective, and we make agent runs for 100 episodes for the purpose of demonstrating the concept. In addition, the average control overhead obtained by the DQN agent is close to where the goal was set. Moreover, the DQN-based agent can even find the parameter set that outperforms the predefined objective.

## 4.4 Performance Comparison between the RL Agents and MBF

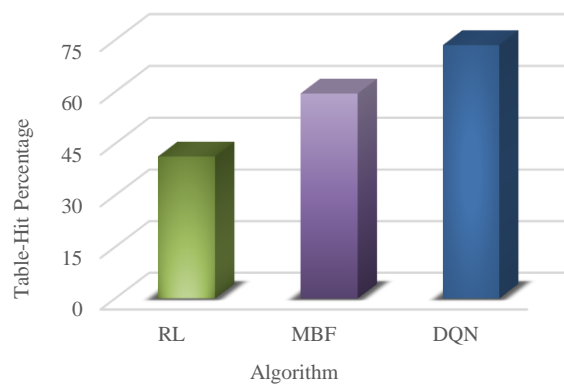

Fig. 10. Performance comparison between our proposed RL agents and the MBF-based approach presented in the recent literature.

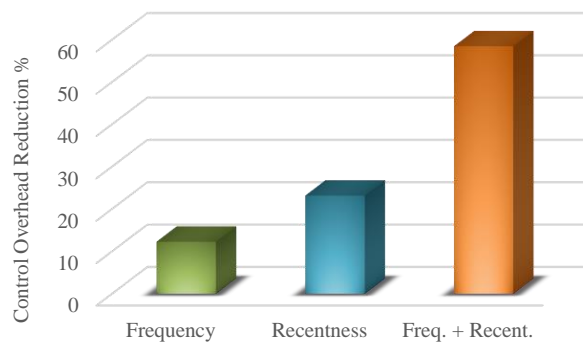

Fig. 11. Significance of each parameter and the significance when they are used jointly.

Next, we performed an experiment to compare the performance of our proposed RL agents with that of the MBF-based approach presented in the recent literature [9], as it utilizes similar parameters in our approach and it is the first work to utilize MBF over flow entry eviction mechanism in OpenFlow switches. Our comparison is conducted based on the flow table-hit ratio. To this end, we utilize the statistics events handler provided by the POX API to collect flow statistics using the *FlowStatsReceived* event. We also generate a traffic pattern with around 8K of active flows and 48 million of packets for the emulation experiment. We use a fixed flow table size of 4K in this experiment. This ensures that the agent does not select flows that exceed this limit.

As shown in Figure 10, the traditional RL agent achieved a table-hit ratio of up to 41.3%, while the MBF-based approach and the DQN agent achieved up to 59.5% [9] and; 73.5% table-hit ratios, respectively. Therefore, our DQN-based approach outperforms the MBF-based approach by 14% in terms of the table-hit ratio, whereas the traditional RL agent achieved 16.4% less table-hit ratio compared to the MBF-based approach.

Therefore, we can conclude on the basis of our experiments that the performance of the DQN-based agent is better than that of the traditional RL agent and the MBF-based approach proposed in [9].

### 4.5 Significance of the Parameters

In our last experiment, we analyzed the significance of each decision parameter for achieving the objective set for the emulation experiment. The results of this experiment are illustrated in Figure 11. When the DQN-based agent is trained with 20 samples with the goal of achieving 60% control plane overhead reduction, the overall control plane overhead reduction was 12.3% considering the flow match frequency parameter alone in the forwarding rules selection process. The overall control plane overhead reduction increased to 23.2% considering the flow recentness parameter alone. This provides the insight that the flow recentness parameter more significantly impacts the performance of the agents compared to the flow match frequency parameter. From Figure 11, it can be seen that the DQN-based agent achieves 58.5% overall control plane overhead reduction when considering both parameters jointly. Thus, our experiments provide the insight that both parameters should be considered in the forwarding rules selection process.

## 5 CONCLUSIONS

In this paper, we present a novel reinforcement learning (RL) approach that automatically finds the values of the decision parameters that can be used to select candidate rules to be placed in the switch's flow table with the objective of minimizing the control plane overhead and enhancing the flow table-hit rate. This machine learning based approach is introduced to minimize the SDN network control plane overhead. Our experiments demonstrate that the proposed RL approach is more effective in minimizing the control plane overhead compared to the MBF approach presented in the recent literature. Performance evaluations confirm that the DQN-based approach outperforms the traditional RL agent and results in faster convergence. Quantitatively, our emulation results illustrate that a reduction of around 60% on the control plane overhead can be achieved using the DQN method with a 13.9% more table-hit rate compared to the MBF method. Our novel approach facilitates an on-line and traffic pattern aware flow entry selection mechanism that effectively improves the utilizing of the switch's TCAM module while reducing the control plane overhead.

Despite the fact that the machine learning based approach performs well in our experiments, further development is needed expand the capabilities of our emulation scope. Currently, our emulation design is based on a simplified network topology and real-world deployments are important to demonstrate the value of the proposed approach. Since such support will enhance the value of the research and enable us to explore the SDN flow management more thoroughly, implementing the support of the known SDN topologies is the focus of our ongoing research.

## REFERENCES

[1] M. Casado, M. J. Freedman, J. Pettit, J. Lou, N. McKeown, and S. Shenker. "Ethane: Taking control of the enterprise." *ACM SIGCOMM Computer Communication Review* 37.4, pp. 1-12, October 2007
[2] N. McKeown, T. Anderson, H. Balakrishnan, G. Parulkar, L. Peterson, J. Rexford, S. Shenker, and J. Turner, "OpenFlow: Enabling Innovation in Campus Networks," in *ACM SIGCOMM Computer Communications Review*, vol. 38, no.2, pp. 73-78, April 2008
[3] Open Networking Foundation. [Online]. Available: https://www.opennetworking.org, February 2017
[4] S. Kandula, S. Sengupta, A. Greenberg, P. Patel, and R. Chaiken, "The nature of data center traffic: Measurement & Analysis," in *Proceedings of the 9th ACM SIGCOMM conference on International measurement*, ser. IMC '09. New York, NY, USA: ACM, 2009, pp. 202-208
[5] T. Benson, A. Akella, and D. A. Maltz, "Network traffic characteristics of data centers in the wild," in *Proceedings of the 10th ACM SIGCOMM conference on Internet measurement*, 2010, pp. 267-280
[6] B. S. Lee, R. Kanagavelu and K. M. M. Aung, "An efficient flow cache algorithm with improved fairness in Software-Defined Data Center Networks," *2013 IEEE 2nd International Conference on Cloud Networking (CloudNet)*, San Francisco, CA, 2013, pp. 18-24
[7] B. Lantz, B. Heller, and N. McKeown, "A Network in a Laptop: Rapid Prototyping for Software-Defined Networks," in *Proceedings of the 9th ACM SIGCOMM Workshop on Hot Topics in Networks*, 2010, p. 19
[8] Mininet – An Instant Virtual Network on your Laptop (or other PC). [Online]. Available: http://www.mininet.org/, February 2017
[9] R. Challa, Y. Lee and H. Choo, "Intelligent eviction strategy for efficient flow table management in OpenFlow Switches," *2016 IEEE NetSoft Conference and Workshops (NetSoft)*, Seoul, 2016, pp. 312-318
[10] "TCAMs and OpenFlow – What Every Practitioner Must Know," [Online]. Available: https://www.sdxcentral.com/articles/contributed/sdn-openflow-tcam-need-to-know/2012/07/, 2012
[11] N. Katta, O. Alipourfard, J. Rexford, and D. Walker, "Inifinite CacheFlow in Software-Defined Networks," in *Proceedings of the 3rd workshop on Hot topics in software defined networking*. ACM, pp. 175-180, 2014
[12] S. Banerjee and K. Kannan, "Tag-In-Tag: Efficient flow table management in SDN switches," *10th International Conference on Network and Service Management (CNSM) and Workshop*, Rio de Janeiro, 2014, pp. 109-117
[13] S. Veeramani, M. Kumar and S. N. Mahammad, "Minimization of flow table for TCAM based openflow switches by virtual compression approach," *2013 IEEE International Conference on Advanced Networks and Telecommunications Systems (ANTS)*, Kattankulathur, 2013, pp. 1-4
[14] H. Zhu, M. Xu, Q. Li, J. Li, Y. Yang and S. Li, "MDTC: An efficient approach to TCAM-based multidimensional table compression," *2015 IFIP Networking Conference (IFIP Networking)*, Toulouse, 2015, pp. 1-9
[15] S. Luo, H. Yu and L. M. Li, "Fast incremental flow table aggregation in SDN," *2014 23rd International Conference on Computer Communication and Networks (ICCCN)*, Shanghai, 2014, pp. 1-8
[16] M. Anan, A. Al-Fuqaha, N. Nasser, T. Mu, and H. Bustam, "Empowering Networking Research and Experimentation through Software-Defined Networking," *Journal of Network and Computer Applications*, vol. 70, pp. 140-155, July 2016
[17] M. Lu, W. Deng and Y. Shi, "TF-IdleTimeout: Improving efficiency of TCAM in SDN by dynamically adjusting flow entry lifecycle," *2016 IEEE International Conference on Systems, Man, and Cybernetics (SMC)*, Budapest, Hungary, 2016, pp. 002681-002686
[18] X. N. Nguyen, D. Saucez, C. Barakat and T. Turletti, "Rules Placement Problem in OpenFlow Networks: A Survey," in *IEEE Communications Surveys & Tutorials*, vol. 18, no. 2, pp. 1273-1286, May 20 2016
[19] R. S. Sutton and A. G. Barto. *Reinforcement Learning: An Introduction*. MIT Press, 1998
[20] S. Marsland. *Machine Learning*. CRC Press, 2nd Edition, 2014
[21] V. Mnih, K. Kavukcuoglu, D. Silver, A. Graves, I. Antonoglou, D. Wierstra, and M. Riedmiller, "Playing Atari with Deep Reinforcement Learning," *arXiv preprint*, arXiv: 1312.5602. December 19, 2013
[22] V. Mnih, K. Kavukcuoglu, D. Silver, AA. Rusu, J. Veness, MG. Bellemare, A. Graves, M. Riedmiller, AK. Fidjeland, G. Ostrovski, and S. Petersen, "Human-level Control through Deep Reinforcement Learning," *Nature*, 518(7540), pages 529-533. February 26, 2015
[23] Google DeepMind. [Online]. Available: https://deepmind.com/, February 2017
[24] J. Schmidhuber, "Deep Learning in Neural Networks: An Overview," *Neural Networks*, vol. 61, pp. 85-117, January 2015
[25] Y. Jia, E. Shelhamer, J. Donahue, S. Karayev, J. Long, R. Girshick, S. Guadarrama, and T. Darrell, "Caffe: Convolutional Architecture for Fast Feature Embedding," in *Proceedings of the 22nd ACM International Conference on Multimedia*, pages 675-678. November 03, 2014
[26] C. Lee, Y. Nakagawa, K. Hyoudou, S. Kobayashi, O. Shiraki and T. Shimizu, "Flow-Aware Congestion Control to Improve Throughput under TCP Incast in Datacenter Networks," 2015 IEEE 39th Annual Computer Software and Applications Conference, Taichung, 2015, pp. 155-162
[27] POX Controller, [Online]. Available: https://github.com/noxrepo/pox
[28] Open vSwitch: An Open Virtual Switch. [Online]. Available: http://www.openvswitch.org, February 2017
[29] H. A. A. Al-Rawi, M. A. Ng, K.-L. A. Yau, "Application of reinforcement learning to routing in distributed wireless networks: a review", *Artificial Intelligence Review*, vol. 43, no. 3, pp. 381-416, 2015
[30] R. Desai, B.P. Patil, "Cooperative reinforcement learning approach for routing in ad hoc networks", in *Proceedings of the 2015 International Conference on Pervasive Computing* (IEEE ICPC 2015), pp. 1-5, January 8–10, 2015
[31] J. Solanki, A. Chauhan, "A Reinforcement Learning Network based Novel Adaptive Routing Algorithm for Wireless Ad-Hoc Network", *International Journal of Science Technology & Engineering.*, vol. 1, no. 12, pp. 135-142, 2015
[32] S. C. Lin, I. F. Akyildiz, P. Wang and M. Luo, "QoS-Aware Adaptive Routing in Multi-layer Hierarchical Software Defined Networks: A Reinforcement Learning Approach," *2016 IEEE International Conference on Services Computing* (SCC), San Francisco, CA, 2016, pp. 25-33
[33] S. Wang, H. Liu, P. H. Gomes, and B. Krishnamachari, "Deep reinforcement learning for dynamic multichannel access", in *International Conference on Computing, Networking and Communications* (ICNC), 2017